\documentclass[10pt,twocolumn,letterpaper]{article}

\usepackage{cvpr}
\usepackage{times}
\usepackage{epsfig}
\usepackage{graphicx}
\usepackage{amsmath}
\usepackage{amssymb}
\usepackage{mathrsfs}
\usepackage{algorithm}
\usepackage{algpseudocode}
\usepackage[pagebackref=true,breaklinks=true,letterpaper=true,colorlinks,bookmarks=false]{hyperref}

\cvprfinalcopy 

\def\cvprPaperID{2248} 

\ifcvprfinal\fi
\begin{document}

\title{Multi-person Articulated Tracking with Spatial and Temporal Embeddings}

\author{
	Sheng Jin$^{1}$ \quad Wentao Liu$^{1}$ \quad Wanli Ouyang$^{2, 3}$ \quad
	Chen Qian $^{1}$  \\
	$^{1}$ SenseTime Research \quad
	$^{2}$ The University of Sydney\\
	$^{3}$ SenseTime Computer Vision Research Group, Australia\\
	$^{1}${\tt\small \{jinsheng, qianchen\}@sensetime.com, liuwtwinter@gmail.com \quad
	$^{2}$ \tt\small wanli.ouyang@sydney.edu.au }
}
	
	\maketitle
	
	\begin{abstract}
	We propose a unified framework for multi-person pose estimation and tracking. Our framework consists of two main components,~\ie~SpatialNet and TemporalNet. The SpatialNet accomplishes body part detection and part-level data association in a single frame, while the TemporalNet groups human instances in consecutive frames into trajectories. Specifically, besides body part detection heatmaps, SpatialNet also predicts the Keypoint Embedding (KE) and Spatial Instance Embedding (SIE) for body part association. We model the grouping procedure into a differentiable Pose-Guided Grouping (PGG) module to make the whole part detection and grouping pipeline fully end-to-end trainable. TemporalNet extends spatial grouping of keypoints to temporal grouping of human instances. Given human proposals from two consecutive frames, TemporalNet exploits both appearance features encoded in Human Embedding (HE) and temporally consistent geometric features embodied in Temporal Instance Embedding (TIE) for robust tracking. Extensive experiments demonstrate the effectiveness of our proposed model. Remarkably, we demonstrate substantial improvements over the state-of-the-art pose tracking method from 65.4\% to 71.8\% Multi-Object Tracking Accuracy (MOTA) on the ICCV'17 PoseTrack Dataset. 
	\end{abstract}
	
	\section{Introduction}
	
	Multi-person articulated tracking aims at predicting the body parts of each person and associating them across temporal periods. It has stimulated much research interest because of its importance in various applications such as video understanding and action recognition~\cite{cheron2015pcnn}. In recent years, significant progress has been made in single frame human pose estimation~\cite{cao2016realtime,he2017mask,Insafutdinov2016DeeperCut,papandreou2017towards}. However, multi-person articulated tracking in complex videos remains challenging. Videos may contain a varying number of interacting people with frequent body part occlusion, fast body motion, large pose changes, and scale variation. Camera movement and zooming further pose challenges to this problem. 
	
	\begin{figure}[t]
		\centering
		\includegraphics[width=0.4\textwidth]{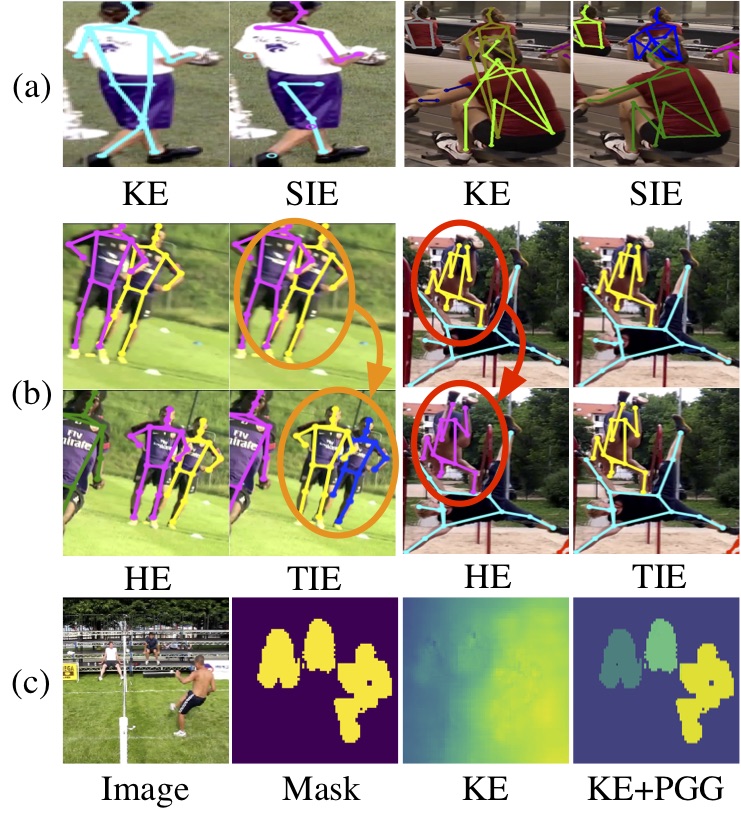}
		\caption{(a) Pose estimation with KE or SIE. SIE may over-segment a single pose into several parts (column 2), while KE may erroneously group far-away body parts together (column 3). (b) Pose tracking with HE or TIE. Poses are color coded by predicted track ids and errors are highlighted by eclipses. TIE is not robust to camera zooming and movement (column 2), while HE is not robust to human pose changes (column 3). (c) Effect of PGG module. Comparing KE before/after PGG (column 3/4), PGG makes embeddings more compact and accurate, where pixels with similar color have higher confidence of belonging to the same person.}
		\label{fig:motivation}
	\end{figure}
	
	Pose tracking~\cite{Iqbal2016PoseTrack} can be viewed as a hierarchical detection and grouping problem. At the part level, body parts are detected and grouped spatially into human instances in each single frame. At the human level, the detected human instances are grouped temporally into trajectories.
	
	Embedding can be viewed as a kind of permutation-invariant instance label to distinguish different instances. Previous works~\cite{newell2017associative} perform keypoint grouping with Keypoint Embedding (KE). KE is a set of 1-D appearance embedding maps where joints of the same person have similar embedding values and those of different people have dissimilar ones. However, due to the over-flexibility of the embedding space, such representations are difficult to interpret and hard to learn~\cite{papandreou2018personlab}. Arguably, a more natural way for the human to assign ids to targets in an image is by counting in a specific order (from left to right and/or from top to bottom). This inspires us to enforce geometric ordering constraints on the embedding space to facilitate training. Specifically, we add six auxiliary ordinal-relation prediction tasks for faster convergence and better interpretation of KE by encoding the knowledge of geometric ordering.
	Recently, Spatial Instance Embedding (SIE)~\cite{nie2017generative,papandreou2018personlab} is introduced for body part grouping. SIE is a 2-D embedding map, where each pixel is encoded with the predicted human center location (x, y). Fig.~\ref{fig:motivation}(a) illustrates the typical error patterns of pose estimation with KE or SIE. SIE may over-segment a single pose into several parts (column 2), while KE sometimes erroneously groups far-away body parts together (column 3). KE better preserves intra-class consistency but has difficulty in separating instances for lack of geometric constraints. Since KE captures appearance features while SIE extracts geometric information, they are naturally complementary to each other. Therefore we combine them to achieve better grouping results.
	
	In this paper, we propose to extend the idea of using appearance and geometric information in a single frame to the temporal grouping of human instances for pose tracking. Previous pose tracking algorithms mostly rely on task-agnostic similarity metrics such as the Object Keypoint Similarity (OKS)~\cite{xiao2018simple,xiu2018pose} and Intersection over Union (IoU)~\cite{girdhar2017detect}. However, such simple geometric cues are not robust to fast body motion, pose changes, camera movement and zoom. For robust pose tracking, we extend the idea of part-level spatial grouping to human-level temporal grouping. Specifically, we extend KE to Human Embedding (HE) for capturing holistic appearance features and extend SIE to Temporal Instance Embedding (TIE) for achieving temporal consistency. Intuitively, appearance features encoded by HE are more robust to fast motion, camera movement and zoom, while temporal information embodied in TIE is more robust to body pose changes and occlusion. We propose a novel TemporalNet to enjoy the best of both worlds. Fig. 1(b) demonstrates typical error patterns of pose tracking with HE or TIE. HE exploits scale-invariant appearance features which are robust to camera zooming and movement (column 1), and TIE preserves temporal consistency which is robust to human pose changes (column 4).
	
	Bottom-up pose estimation methods follow the two-stage pipeline to generate body part proposals at the first stage and group them into individuals at the second stage. Since the grouping is mainly used as post-processing, \ie graph based optimization~\cite{Insafutdinov2016ArtTrack,Insafutdinov2016DeeperCut,Iqbal2016PoseTrack,jin2017towards,pishchulin2016deepcut} or heuristic parsing~\cite{cao2016realtime,papandreou2018personlab}, no error signals from the grouping results are back-propagated. We instead propose a fully differentiable Pose-Guided Grouping (PGG) module, making detection-grouping fully end-to-end trainable. We are able to directly supervise the grouping results and the grouping loss is back-propagated to the low-level feature learning stages. This enables more effective feature learning by paying more attention to the mistakenly grouped body parts. Moreover, to obtain accurate regression results, post-processing clustering~\cite{nie2017generative} or extra refinement~\cite{papandreou2018personlab} are required. Our PGG helps to produce accurate embeddings (see Fig.~\ref{fig:motivation}(c)). To improve the pose tracking accuracy, we further extend PGG to temporal grouping of TIE.
	
	In this work, we aim at unifying pose estimation and tracking in a single framework. SpatialNet detects body parts in a single frame and performs part-level spatial grouping to obtain body poses. TemporalNet accomplishes human-level temporal grouping in consecutive frames to track targets across time. These two modules share the feature extraction layers to make more efficient inference.
	
	The main contributions are summarized as follows: 
	\begin{itemize}
		\item For pose tracking, we extend the KE and SIE in still images to Human Embedding (HE) and Temporal Instance Embeddings (TIE) in videos. HE captures human-level global appearance features to avoid drifting in camera motion, while TIE provides smoother geometric features to obtain temporal consistency. 
		
		\item A fully differentiable Pose-Guided Grouping (PGG) module for both pose estimation and tracking, which enables the detection and grouping to be fully end-to-end trainable. The introduction of PGG and its grouping loss significantly improves the spatial/temporal embedding prediction accuracy.
	\end{itemize}
	
	\section{Related Work}
	\label{sec:related_work}
	
	\subsection{Multi-person Pose Estimation in Images}
	Recent multi-person pose estimation approaches can be classified into top-down and bottom-up methods.
	\textbf{Top-down} methods~\cite{fang2016rmpe,he2017mask,xiao2018simple,papandreou2017towards} locate each person with a bounding box then apply single-person pose estimation. They mainly differ in the choices of human detectors~\cite{renNIPS15fasterrcnn} and single-person pose estimators~\cite{newell2016stacked,wei2016convolutional}. They highly rely on the object detector and may fail in cluttered scenes, occlusion, person-to-person interaction, or rare poses. More importantly, top-down methods perform single-person pose estimation individually for each human candidate. Thus, its inference time is proportional to the number of people, making it hard for achieving real-time performance. Additionally, the interface between human detection and pose estimation is non-differentiable, making it difficult to train in an end-to-end manner. 
	\textbf{Bottom-up} approaches~\cite{cao2016realtime,Insafutdinov2016DeeperCut,pishchulin2016deepcut} detect body part candidates and group them into individuals. Graph-cut based methods~\cite{Insafutdinov2016DeeperCut,pishchulin2016deepcut} formulate grouping as solving a graph partitioning based optimization problem, while ~\cite{cao2016realtime,papandreou2018personlab} utilize the heuristic greedy parsing algorithm to speed up decoding. However, these bottom-up approaches only use grouping as post-processing and no error signals from grouping results are back-propagated. 
	
	More recently, efforts have been devoted to end-to-end training or joint optimization. For top-down methods, Xie \etal~\cite{xie2018environment} proposes a reinforcement learning agent to bridge the object detector and the pose estimator. For bottom-up methods, Newell \etal~\cite{newell2017associative} proposes the keypoint embedding (KE) to tag instances and train by pairwise losses. Our framework is a bottom-up method inspired by \cite{newell2017associative}. \cite{newell2017associative} supervises the grouping in an indirect way. It trains keypoint embedding descriptors to ease the post-processing grouping. However, no direct supervision on grouping results is provided. Even if the pairwise loss of KE is low, it is still possible to produce wrong grouping results, but \cite{newell2017associative} does not model such grouping loss. We instead propose a differentiable Pose-Guided Grouping (PGG) module to learn to group body parts, making the whole pipeline fully end-to-end trainable, yielding significant improvement in pose estimation and tracking.
	
	Our work is also related to~\cite{nie2017generative,papandreou2018personlab}, where spatial instance embeddings (SIE) are introduced to aid body part grouping. However, due to lack of grouping supervision, their embeddings are always noisy~\cite{nie2017generative,papandreou2018personlab} and additional clustering~\cite{nie2017generative} or refinement~\cite{papandreou2018personlab} is required. We instead employ PGG and additional grouping losses to learn to group SIE, making it end-to-end trainable while resulting in much more compact embedding representation.
	
	\begin{figure}
		\centering
		\includegraphics[width=0.45\textwidth]{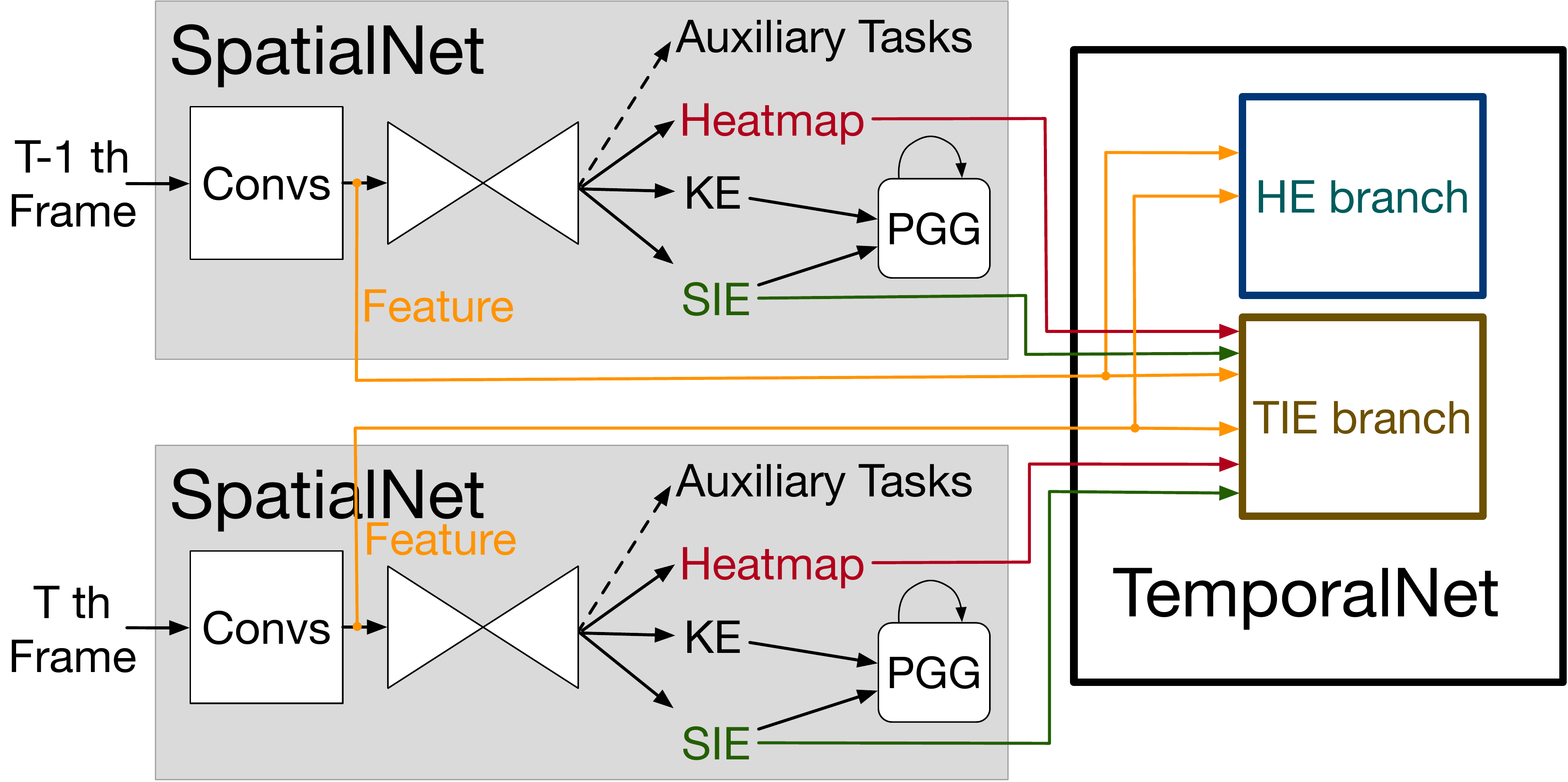}
		\caption{The overview of our framework for pose tracking.}
		\label{fig:framework}
	\end{figure}
	
	\subsection{Multi-person Pose Tracking}
	
	Recent works on multi-person pose tracking mostly follow the tracking-by-detection paradigm, in which human body parts are first detected in each frame, then data association is performed over time to form trajectories. 
	
	Offline pose tracking methods take future frames into consideration, allowing for more robust predictions but having high computational complexity. ProTracker~\cite{girdhar2017detect} employs 3D Mask R-CNN to improve the estimation of body parts by leveraging temporal context encoded within a sliding temporal window. Graph partitioning based methods~\cite{Insafutdinov2016ArtTrack,Iqbal2016PoseTrack,jin2017towards} formulate multi-person pose tracking into an integer linear programming (ILP) problem and solve spatial-temporal grouping. Such methods achieve competitive performance in complex videos by enforcing long-range temporal consistency.
	
	Our approach is an online pose tracking approach, which is faster and fits for practical applications. Online pose tracking methods~\cite{doering2018joint,payer2017simultaneous,zhu2017multi,xiao2018simple} mainly use bi-partite graph matching to assign targets in the current frame to existing trajectories. However, they only consider part-level geometric information and ignore global appearance features. When faced with fast pose motion and camera movement, such geometrical trackers are prone to tracking errors. We propose to extend SpatialNet to TemporalNet to capture both appearance features in HE and temporal coherence in TIE, resulting in much better tracking performance.
	
	\section{Method}
	As demonstrated in Figure~\ref{fig:framework}, we unify pose estimation and tracking in a single framework. Our framework consists of two major components: SpatialNet and TemporalNet. 
	
	SpatialNet tackles multi-person pose estimation by body part detection and part-level spatial grouping. It processes a single frame at a time. Given a frame, SpatialNet produces heatmaps, KE, SIE and geometric-ordinal maps simultaneously. Heatmaps model the body part locations. KE encodes the part-level appearance features, while SIE captures the geometric information about human centers. The auxiliary geometric-ordinal maps enforce ordering constraints on the embedding space to facilitate training of KE. PGG is utilized to make both KE and SIE to be more compact and discriminative. We finally generate the body pose proposals by greedy decoding following~\cite{newell2017associative}.
	
	TemporalNet extends SpatialNet to deal with online human-level temporal grouping. It consists of HE branch and TIE branch, and shares the same low-level feature extraction layers with SpatialNet. Given body pose proposals, HE branch extracts region-specific embedding (HE) for each human instance. TIE branch exploits the temporally coherent geometric embedding (TIE). Given HE and TIE as pairwise potentials, a simple bipartite graph matching problem is solved to generate pose trajectories. 
	
	\subsection{SpatialNet: Part-level Spatial Grouping}
	
	Throughout the paper, we use following notations. Let $p=(x,y) \in \mathbb{R}^2$ be the 2-D position in an image, and $p_{j,k} \in \mathbb{R}^2$ the location of body part $j$ for person $k$. We use $P_k = \{p_{j,k}\}_{j=1:J}$ to represent the body pose of the $k$th person. We use 2D Gaussian confidence heatmaps to model the body part locations. Let $C_{j,k}$ be the confidence heatmap for the $j$th body part of $k$th person, which is calculated by $C_{j,k}(p)=\exp(-\|p-p_{j,k}\|_2^2/\sigma^2)$ for each position $p$ in the image, where $\sigma$ is set as 2 in the experiments. Following~\cite{cao2016realtime}, we take the maximum of the confidence heatmaps to get the ground truth confidence heatmap, \ie $C^*_{j}(p)=\max_kC^*_{j,k}(p)$. 
	
	The detection loss is calculated by weighted $\ell_2$ distance respect to the ground truth confidence heatmaps.
	\begin{equation}
	L_{det} = \sum_j\sum_p \|C^*_{j}(p)-C_{j}(p)\|_2^2.
	\end{equation}
	
	\subsubsection{Keypoint Embedding (KE) with auxiliary tasks}
	\label{sec:KE}
	
	We follow~\cite{newell2017associative} to produce the keypoint embedding $\mathcal{K}$ for each type of body part. However, such kind of embedding representation has several drawbacks. First, the embedding is difficult to interpret~\cite{newell2017associative,papandreou2018personlab}. Second, it is hard to learn due to its over-flexibility with no direct supervision available. To overcome these drawbacks, we introduce several auxiliary tasks to facilitate training and improve interpretation. The idea of auxiliary learning~\cite{suddarth1990rule} has shown effective both in supervised learning~\cite{qi2016volumetric} and reinforcement learning~\cite{jaderberg2016reinforcement}. Here, we explore auxiliary training in the context of keypoint embedding representation learning. 
	
	By auxiliary training, we explicitly enforce the embedding maps to learn geometric ordinal relations. Specifically, we define six auxiliary tasks: to predict the 'left-to-right' \textit{l2r}, 'right-to-left' \textit{r2l}, 'top-to-bottom' \textit{t2b}, 'bottom-to-top' \textit{b2t}, 'far-to-near' \textit{f2n} and 'near-to-far' \textit{n2f} orders of human instances in a single image. For example, in the `left-to-right' map, the person from left to right in the images should have low to high order (value). 
	Fig.~\ref{fig:visual} (c)(d)(e) visualize some example predictions of the auxiliary tasks. We see human instances are clearly arranged in the corresponding geometric ordering. We also observe that KE (Fig.~\ref{fig:visual} (b)) and the geometric ordinal-relation maps (c)(d)(e) share some similar patterns, which suggests that KE acquires some knowledge of geometric ordering. 
	
	Following~\cite{newell2017associative}, $\mathcal{K}$ is trained with pairwise grouping loss $L_{KE} = L_{pull}+L_{push}$. The pull loss (Eq.~\ref{eq:pull_loss}) is computed as the squared distance between the human reference embedding and the predicted embedding of each joint. The push loss (Eq.~\ref{eq:push_loss}) is calculated between different reference embeddings, which exponentially drops to zero as the increase of embedding difference. Formally, we define the reference embedding for the $k$th person as $\bar{m}_{\cdot, k} = \frac{1}{J} \sum_j m_j(p_{j,k})$. 
	
	\begin{equation}
	\label{eq:pull_loss}
	L_{pull}= \frac{1}{J\cdot K} \sum_k \sum_j \|m(p_{j,k}) - \bar{m}_{\cdot, k} \|_2^2.
	\end{equation}
	\begin{equation}
	\label{eq:push_loss}
	L_{push}= \frac{1}{K^2} \sum_k \sum_{k'}  \exp \{-\frac{1}{2}(\bar{m}_{\cdot, {k}}- \bar{m}_{\cdot, {k'}})^2 \}.
	\end{equation}
	
	For auxiliary training, we replace the push loss with the ordinal loss but keep the pull loss (Eq.~\ref{eq:pull_loss}) the same.
	\begin{align}
	L_{aux} &= \frac{1}{K^2} \sum_k \sum_{k'} \log(1 + \exp(Ord * (\bar{m}_{\cdot, k}- \bar{m}_{\cdot, {k'}}) )) \nonumber \\
	&+\frac{1}{J\cdot K} \sum_k \sum_j \|m(p_{j,k}) - \bar{m}_{\cdot, k} \|_2^2,
	\end{align}
	where $Ord = \{1,-1\}$ indicates the ground-truth order for person $k$ and $k'$. In \textit{l2r}, \textit{r2l}, \textit{t2b}, and \textit{b2t}, we sort human instances by their centroid locations. For example, in \textit{l2r} , if $k$th person is on the left of $k'$th person, then $Ord=1$, otherwise $Ord=-1$. In \textit{f2n} and \textit{n2f}, we sort them according to the head size $\|p_{headtop, k} - p_{neck, k}\|_2^2$.
	
	\subsubsection{Spatial Instance Embedding (SIE)}
	\label{sec:SIE}
	
\begin{figure}
		\centering
		\includegraphics[width=0.49\textwidth]{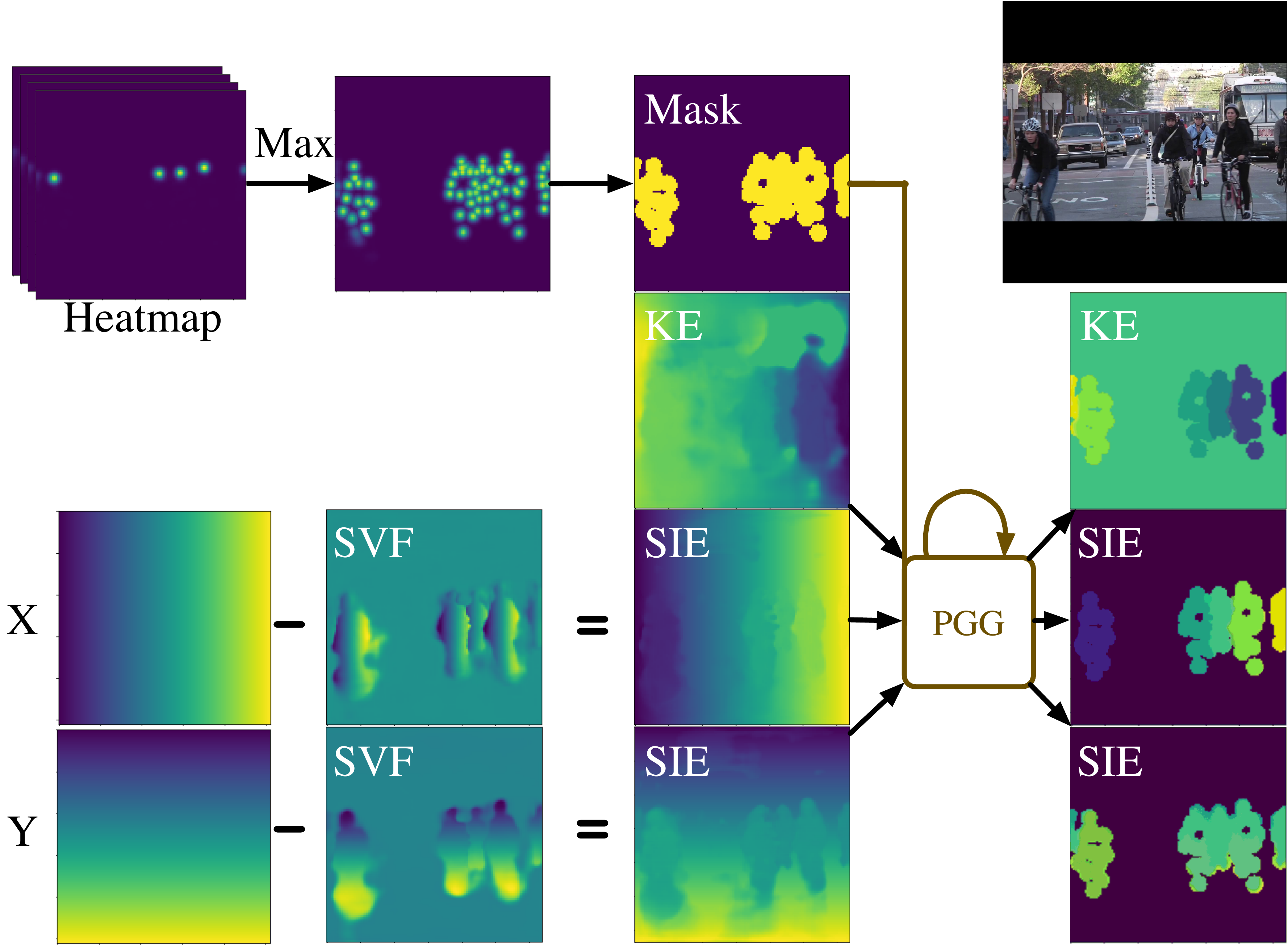}
		\caption{Spatial keypoint grouping with Pose-Guided Grouping (PGG). We obtain more compact and accurate Keypoint Embedding (KE) and Spatial Instance Embedding (SIE) with PGG.}
		\label{fig:pgg}
	\end{figure}

\begin{figure*}
		\centering
		\includegraphics[width=0.9\textwidth]{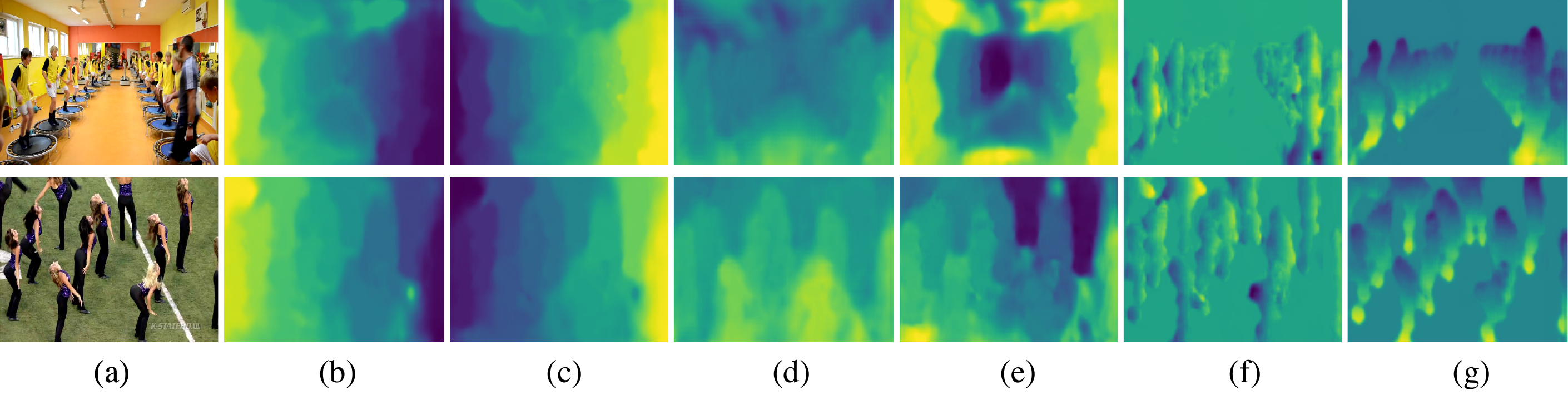}
		\caption{(a) input image. (b) the average KE. (c)(d)(e) predicted 'left-to-right', 'top-to-bottom' and 'far-to-near' geometric-relation maps. We use colors to indicate the predicted orders, where the brighter color means the higher ordinal value. (f)(g) are the spatial vector fields of x-axis and y-axis respectively. The bright color means positive offset relative to the human center, while dark color means negative.}
		\vspace{-0.4cm}
		\label{fig:visual}
	\end{figure*}

	For lack of geometric information, KE has difficulty in separating instances and tends to erroneously group with distant body parts. To remedy this, we combine KE with SIE to embody instance-wise geometric cues. Concretely, we predict the dense offset spatial vector fields (SVF), where each 2-D vector encodes the relative displacement from the human center to its absolute location $p$. Fig.~\ref{fig:visual}(f)(g) visualize the spatial vector fields of x-axis and y-axis, which distinguish the left/right sides and upper/lower sides relative to its body center. As shown in Fig.~\ref{fig:pgg}, subtracted by its coordinate, SVF can be decoded to SIE in which each pixel is encoded with the human center location.
	
	We denote the spatial vector fields (SVF) by $\hat{\mathcal{S}}$, and SIE by $\mathcal{S}$. We use $\ell_1$ distance to train SVF, where the ground truth spatial vector is the displacement from the person center to each body part. 
	
	\begin{equation}
	\label{eq:loss_svf}
	L_{SIE} = \frac{1}{J\cdot K}\sum_{j=1}^{J}\sum_{k=1}^K \|\hat{\mathcal{S}}(p_{j,k}) -  (p_{j,k} - p_{\cdot,k})\|_1,
	\end{equation}
	where $p_{\cdot,k} = \frac{1}{J} \sum_j p_{j,k}$, is the center of person $k$. 
	
	\subsection{Pose-Guided Grouping (PGG) Module}
	\vspace{-0.3cm}
	~\label{sec:pgg}
	
	In prior bottom-up methods~\cite{cao2016realtime,nie2017generative,papandreou2018personlab}, detection and grouping are separated. We reformulate the grouping process into a differentiable Pose-Guided Grouping (PGG) module for end-to-end training. By directly supervising the grouping results, more accurate estimation is obtained.
	
	Our PGG is based on Gaussian Blurring Mean Shift (GBMS)~\cite{carreiraperpinan2008generalised} algorithm and inspired by~\cite{kong2017recurrent}, which is originally proposed for segmentation. However, directly applying GBMS in the challenging articulate tracking task is not desirable. First, the complexity of GBMS is $O(n^2)$, where $n$ is the number of feature vectors to group. Direct use of GBMS on the whole image will lead to huge memory consumption. Second, the predicted embeddings are always noisy especially in background regions, where no supervision is available during training. As illustrated in the top row of Fig.~\ref{fig:visual}, embedding noises exist in the background area (the ceiling or the floor). The noise in these irrelevant regions will affect the mean-shift grouping accuracy. We propose a novel Pose-Guided Grouping module to address the above drawbacks. Considering the sparseness of the matrix (body parts only occupy a small area in images), we propose to use the human pose mask to guide grouping, which rules out irrelevant areas and significantly reduces the memory cost. As shown in Fig.~\ref{fig:pgg}, we apply \emph{max} along the channel $\bar{C}(p) = \max_j C_j(p)$ and generate the instance-agnostic pose mask $\mathbf{M} \in \mathbb{R}^{W \times H} $, by thresholding at $\tau=0.2$. $\mathbf{M}(p)$ is 1 if $\bar{C}(p)>\tau$, otherwise 0.
	
	\renewcommand{\algorithmicrequire}{\textbf{Input:}}
	\renewcommand{\algorithmicensure}{\textbf{Output:}}
	\begin{algorithm}[tb]
		\caption{Pose-Guided Grouping} \label{alg:pgg}
		\begin{algorithmic}[1]
			\Require
			KE $\mathcal{K}$, SIE $\mathcal{S}$, Mask $\mathbf{M}$, and iteration number $R$.
			\Ensure
			$\mathcal{X}$
			\State Concatenate $\mathcal{K}$ and $\mathcal{S}$, mask-selected by $\mathbf{M}$, and reshape to $\mathbf{X}^{(1)} \in \mathbb{R}^{D \times N}$.
			\State Initialize $\mathcal{X} = \left[ \mathbf{X}^{(1)} \right]$
			\For{$r=1,2,\cdots R$}
			\State Gaussian Affinity $\mathbf{W}^{(r)} \in \mathbb{R}^{N \times N}$.
			$\mathbf{W}^{(r)}(i,j) = \exp(-\frac{\delta^2}{2}  \Arrowvert x_i^{(r)}-x_j^{(r)} \Arrowvert^2_2)$, $\forall x_i^{(r)}, x_j^{(r)} \in \mathbf{X}^{(r)}$.
			\State Normalization Matrix. $\mathbf{D}^{(r)} = diag \big(\mathbf{W}^{(r)} \cdot \vec{\mathbf{1}} \big)$
			\State Update. $\mathbf{X}^{(r+1)} = \mathbf{X}^{(r)}\mathbf{W}^{(r)} \big(\mathbf{D}^{(r)} \big)^{-1}$
			\State $\mathcal{X} = \left[\mathcal{X} ; \mathbf{X}^{(r+1)} \right]$
			\EndFor\\
			\Return
			$\mathcal{X}$
		\end{algorithmic}
	\end{algorithm}
	
	Both spatial (KE and SIE) and temporal (TIE) embeddings can be grouped by PGG. Take spatial grouping for example, we refine KE and SIE with PGG module to get more compact and discriminative embedding descriptors. The Pose-Guided Grouping algorithm is summarized in Alg.~\ref{alg:pgg}. KE and SIE are first concatenated to $D \times W \times H$ dimensional feature maps. Then embeddings are selected according to the binary pose mask $\mathbf{M}$ and reshaped to $\mathbf{X}^{(1)} \in \mathbb{R}^{D \times N} $ as initialization, where $N$ is the number of non-zero elements in $\mathbf{M}$, ($N \ll W \times H$). Recurrent mean-shift grouping is then applied to $\mathbf{X}^{(1)}$ for $R$ iterations. In each iteration, the Gaussian affinity is first calculated with the isotropic multivariate normal kernel $\mathbf{W} = \exp(-\frac{\delta^2}{2}  \Arrowvert x-x_i \Arrowvert^2_2)$, where the kernel bandwidth $\delta$ is empirically chosen as 5 in the experiments. $\mathbf{W} \in \mathbb{R}^{N \times N}$ can be viewed as the weighted adjacency matrix. The diagonal matrix of affinity row sum $\mathbf{D} = diag(\mathbf{W} \cdot \vec{\mathbf{1}})$ is used for normalization, where $\vec{\mathbf{1}}$ means a vector with all entries one. We then update $\mathbf{X}$ with the normalized Gaussian kernel weighted mean, $\mathbf{X} = \mathbf{X}\mathbf{W}\mathbf{D}^{-1}$. After several iterations of grouping refinement, the embeddings become distinct for heterogeneous pairs and similar for homogeneous ones. When training, we apply the pairwise pull/push losses (Eq.~\ref{eq:pull_loss} and~\ref{eq:push_loss}) over all iterations of grouping results $\mathcal{X}$.
	
	\subsection{TemporalNet: Human Temporal Grouping}
	\label{sec:TemporalNet}
	
	TemporalNet extends SpatialNet to perform human-level temporal grouping in an online manner. Formally, we use the superscript $t$ to distinguish different frames. $I^t$ denotes the input frame at time-step $t$, which contains $K^t$ persons. SpatialNet is applied to $I^t$ to estimate a set of poses $\mathcal{P}^t = \lbrace {P}^{t}_1, \ldots {P}^{t}_{K^t} \rbrace$. TemporalNet aims at temporally grouping human pose proposals $\mathcal{P}^t$ in the current frame with already tracked poses $\mathcal{P}^{t-1}$ in the previous frame. TemporalNet exploits both human-level appearance features (HE) and temporally coherent geometric information (TIE) to calculate the total pose similarity. Finally, we generate the pose trajectories by solving the bipartite graph matching problems, using pose similarity as pairwise potentials.
	\vspace{-0.2cm}
	\subsubsection{Human Embedding (HE)}
	\label{sec:HE}
	
	To obtain human-level appearance embedding (HE), we introduce a region-specific HE branch based on~\cite{yu2017devil}. Given predicted pose proposals, HE brach first calculates human bounding boxes to cover the corresponding human keypoints. For each bounding box, ROI-Align pooling~\cite{he2017mask} is applied to the shared low-level feature maps to extract region-adapted ROI features. The ROI features are then mapped to the human embedding $\mathcal{H} \in \mathbb{R}^{3072}$. HE is trained with triplet loss~\cite{schroff2015facenet}, pulling HE of the same instance closer, and pushing apart embeddings of different instances.
	\begin{equation}
	L_{HE}=\sum_{\substack{k_1=k_2\\   k_1\neq k_3}} \max(0,\| \mathcal{H}_{k_1} - \mathcal{H}_{k_2} \|_2^2-\| \mathcal{H}_{k_1} - \mathcal{H}_{k_3} \|_2^2+\alpha),
	\end{equation}
	where the margin term $\alpha$ is set to 0.3 in the experiments.

	\subsubsection{Temporal Instance Embedding (TIE)}
	\label{sec:TIE}
	To exploit the temporal information for pose tracking, we naturally extend the Spatial Instance Embedding (SIE) to the Temporal Instance Embedding (TIE). TIE branch concatenates low-level features, body part detection heatmaps and SIE from two neighboring frames. The concatenated feature maps are then mapped to dense TIE. 
	
	TIE is a task-specific representation which measures the displacement between the keypoint of one frame and the human center of another frame. This design utilizes the mutual information between keypoint and human in adjacent frames to handle occlusion and pose motion simultaneously. Specifically, we introduce bi-directional temporal vector fields (TVF), which are denoted as $\hat{\mathcal{T}}$ and $\hat{\mathcal{T}'}$ respectively. Forward TVF $\hat{\mathcal{T}}$ encodes the relative displacement from the human center in $(t-1)$-th frame to body parts in the $t$-th frame, it temporally propagates the human centroid embeddings from $(t-1)$-th to $t$-th
	frame. In contrast, Backward TVF $\hat{\mathcal{T}'}$ represents the offset from current $t$-th frame body center to body parts in the previous frame. 
	
	\begin{align}
	\label{eq:loss_tvf}
	L_{TIE} & =\frac{1}{J\cdot K^{t}}\sum_{j=1}^{J}\sum_{k=1}^{K^{t}} \|\hat{\mathcal{T}}(p^t_{j,k}) - (p_{j,k}^{t} - p_{\cdot,k}^{t-1}) \|_1 \nonumber \\
	+ &\frac{1}{J\cdot K^{t-1}}\sum_{j=1}^{J}\sum_{k'=1}^{K^{t-1}} \|\hat{\mathcal{T}'}(p^{t-1}_{j,k'}) - (p_{j,k'}^{t-1} - p_{\cdot,k'}^t) \|_1,
	\end{align}
	where $p^t_{\cdot,k} = \frac{1}{J} \sum_j p^t_{j,k}$, is the center of person $k$ at time step $t$. Simply subtracted from absolute locations, we get the corresponding Forward TIE $\mathcal{T}$ and Backward TIE $\mathcal{T}'$. Thereby, TIE encodes the temporally propagated human centroid. Likewise, we also extend the idea of spatial grouping to temporal grouping. TemporalNet outputs Forward TIE $\mathcal{T}$ and Backward TIE $\mathcal{T}'$, which are refined by PGG independently. Take Forward TIE $\mathcal{T}$ for example, we generate pose mask $M$ using body heatmaps from the $t$-th frame. We rule out irrelevant regions of $\mathcal{T}$ and reshape it to $\mathbf{X}^{(1)} \in \mathbb{R}^{D \times N}$. Subsequently, recurrent mean-shift grouping is applied. Again, additional grouping losses (Eq.~\ref{eq:pull_loss},\ref{eq:push_loss}) are used to train TIE.

	\subsubsection{Pose Tracking}
	\label{sec:robust_tracking}
	
	The problem of temporal pose association is formulated as a bipartite graph based energy maximization problem. 
	The estimated poses $\mathcal{P}^t$ are then associated with the previous poses $\mathcal{P}^{t-1}$ by bipartite graph matching.
	\begin{align}\label{eq:objective}
	\hat{z} & =\underset{z}{\textit{argmax}} \sum_{P_k^t \in \mathcal{P}^{t}} \sum_{P^{t-1}_{k'} \in \mathcal{P}^{t-1}} \Psi_{P_k^t ,P^{t-1}_{k'}} \cdot z_{P_k^t ,P^{t-1}_{k'}} \\
	&\text{s.t.} \;\;\; \forall P_k^t \in \mathcal{P}^{t}, \; \sum_{P^{t-1}_{k'} \in \mathcal{P}^{t-1}} z_{P_k^t, P^{t-1}_{k'}} \leq 1 
	\nonumber \\ 
	&\text{and} \;\;\;\forall P^{t-1}_{k'} \in \mathcal{P}^{t-1}, \; \sum_{P_k^t \in \mathcal{P}^t} z_{P_k^t, P^{t-1}_{k'}} \leq 1, \nonumber
	\end{align}
	where $z_{P_k^t ,P^{t-1}_{k'}} \in \{0,1\}$ is a binary variable which implies if the pose hypothesis $P_k^t$ and $P^{t-1}_{k'}$ are associated. The pairwise potentials $\Psi$ represent the similarity between pose hypothesis. $\Psi = \lambda_{HE} \Psi_{HE} + \lambda_{TIE} \Psi_{TIE}$, with $\Psi_{HE}$ for human-level appearance similarity and $\Psi_{TIE}$ for temporal smoothness. $\lambda_{HE}$ and $\lambda_{TIE}$ are hyperparameters to balance them, with $\lambda_{HE}=3$ and $\lambda_{TIE}=1$.
	
	The human-level appearance similarity is calculated as the $\ell_2$ embedding distance: $\Psi_{HE} = \| \mathcal{H}_{k} - \mathcal{H}_{k'} \|_2^2.$
	And the temporal smoothness term $\Psi_{TIE}$ is computed as the similarity between the encoded human center locations in SIE $\mathcal{S}$ and the temporally propagated TIE $\mathcal{T}$, $\mathcal{T}'$.
	
	\begin{align}
	\Psi_{TIE} =  \frac{1}{2J} \sum_{j=1}^{J}& \left( \| \mathcal{T}'(p_{j,k'}^{t-1}) - \mathcal{S}^t(p_{j,k}^t)\|_2^2\right. \nonumber \\
	& + \left. \| \mathcal{T}(p_{j,k}^{t}) - \mathcal{S}^{t-1}(p_{j,k'}^{t-1})\|_2^2  \right), 
	\end{align}
	
	The bipartite graph matching problem (Eq.~\ref{eq:objective}) is solved using Munkres algorithm to generate pose trajectories. 
	
	\subsection{Implementation Details}
	
	Following~\cite{newell2017associative}, SpatialNet uses the 4-stage stacked-hourglass as its backbone. We first train SpatialNet without PGG. The total losses consist of $L_{det}, L_{KE}, L_{aux}$ and $L_{SIE}$, with their weights 1:1e-3:1e-4:1e-4. We set the initial learning rate to 2e-4 and reduce it to 1e-5 after 250K iterations. Then we fine-tune SpatialNet with PGG included. In practice, we have found the iteration number $R=1$ is sufficient, and more iterations do not lead to much gain.  
	
	TemporalNet uses 1-stage hourglass model~\cite{newell2016stacked}. When training, we simply fix SpatialNet and train TemporalNet for another 40 epochs with learning rate of 2e-4. We randomly select a pair of images $I^t$ and $I^{t'}$ from a range-5 temporal window ($\|t-t'\|_1 \leq 5$) in a video clip as input.
	
	\section{Experiments}
	\label{sec:exp}
	
	\subsection{Datasets and Evaluation}
	
	\textbf{MS-COCO Dataset}~\cite{lin2014microsoft} contains over 66k images with 150k people and 1.7 million labeled keypoints, for pose estimation in images. For the MS-COCO results, we follow the same train/val split as~\cite{newell2017associative}, where a held-out set of 500 training images are used for evaluation.
	
	\textbf{ICCV'17 PoseTrack Challenge Dataset}~\cite{andriluka2017posetrack} is a large-scale benchmark for multi-person articulated tracking, which contains 250 video clips for training and 50 sequences of videos for validation. 
	
	\textbf{Evaluation Metrics:} We follow~\cite{andriluka2017posetrack} to use AP to evaluate multi-person pose estimation and the multi-object tracking accuracy (MOTA)~\cite{bernardin2008evaluating} to measure tracking performance. 
	
	\subsection{Comparisons with the State-of-the-art Methods}
	
	We compare our framework with the state-of-the-art methods on both pose estimation and tracking on the ICCV'17 PoseTrack validation set. As a common practice~\cite{andriluka2017posetrack}, additional images from MPII-Pose~\cite{andriluka14cvpr} are used for training. Table~\ref{tab:test_kpt} demonstrate our single-frame pose estimation performance. We show that our model achieves the state-of-the-art $77.0$ mAP without single-person pose model refinement. Table~\ref{tab:test_tracking} evaluates the multi-person articulated tracking performance. Our model outperforms the state-of-the-art methods by a large margin. Compared with the winner of ICCV'17 PoseTrack Challenge (ProTracker~\cite{girdhar2017detect}), our method obtain an improvement of 16.6\% in MOTA. Our model further improves over the current state-of-the-art pose tracker (FlowTrack~\cite{xiao2018simple}) by 6.4\% in MOTA with comparable single frame pose estimation accuracy, indicating the effectiveness of our TemporalNet.

	\begin{table}[!htpb] \footnotesize
		\centering
		\scalebox{0.8}{
			\begin{tabular}{l|c|c|c|c|c|c|c|c} \hline
				Method & Head & Shou & Elb  & Wri  & Hip  & Knee & Ankl & Total\\ \hline
				ProTracker~\cite{girdhar2017detect} 
				&$69.6$&$73.6$&$60.0$&$49.1$&$65.6$&$58.3$&$46.0$&$60.9$  \\ 
				PoseFlow~\cite{xiu2018pose} 
				&$66.7$&$73.3$&$68.3$&$61.1$&$67.5$&$67.0$&$61.3$&$66.5$  \\ 
				BUTDS~\cite{jin2017towards} 
				&$79.1$&$77.3$&$69.9$&$58.3$&$66.2$&$63.5$&$54.9$&$67.8$ \\ 
				ArtTrack~\cite{andriluka2017posetrack} 
				&$78.7$&$76.2$&$70.4$&$62.3$&$68.1$&$66.7$&$58.4$&$68.7$  \\ 
				ML\_Lab~\cite{zhu2017multi} 
				&$\mathbf{83.8}$&$\mathbf{84.9}$&$76.2$&$64.0$&$72.2$&$64.5$&$56.6$&$72.6$\\
				FlowTrack~\cite{xiao2018simple} 
				&$81.7$&$83.4$&$\mathbf{80.0}$&$\mathbf{72.4}$&$75.3$&$\mathbf{74.8}$&$67.1$&$76.9$  \\ \hline 
				Ours 
				&$\mathbf{83.8}$&$81.6$&$77.1$&$70.0$&$\mathbf{77.4}$&$74.5$&$\mathbf{70.8}$&$\mathbf{77.0}$ \\ \hline
		\end{tabular}}
		\vspace{0cm}
		\caption{Comparisons with the state-of-the-art methods on single-frame pose estimation on ICCV'17 PoseTrack Challenge Dataset. \label{tab:test_kpt}}
	\end{table}

	\begin{table}[!htpb] \small
		\centering
		\vspace{-0.6cm}
		\scalebox{0.75}{
			\setlength{\tabcolsep}{3pt}
			\begin{tabular}{l|c|c|c|c|c|c|c|c} \hline
				Method& MOTA & MOTA & MOTA & MOTA & MOTA & MOTA & MOTA & MOTA   \\
				& Head & Shou & Elb  & Wri  & Hip  & Knee & Ankl & Total\\ \hline
				ArtTrack~\cite{andriluka2017posetrack} 
				&$66.2$&$64.2$&$53.2$&$43.7$&$53.0$&$51.6$&$41.7$&$53.4$  \\  
				ProTracker~\cite{girdhar2017detect} 
				&$61.7$&$65.5$&$57.3$&$45.7$&$54.3$&$53.1$&$45.7$&$55.2$\\  
				BUTD2~\cite{jin2017towards} 
				&$71.5$&$70.3$&$56.3$&$45.1$&$55.5$&$50.8$&$37.5$&$56.4$\\ 
				PoseFlow~\cite{xiu2018pose}
				&$59.8$&$67.0$&$59.8$&$51.6$&$60.0$&$58.4$&$50.5$&$58.3$\\  
				JointFlow~\cite{doering2018joint} & - & - & - &  - & - & - & - & $59.8$  \\
				FlowTrack~\cite{xiao2018simple} &$73.9$&$75.9$&$63.7$&$56.1$&$65.5$&$65.1$&$53.5$&$65.4$ \\\hline
				Ours 
				&$\mathbf{78.7}$&$\mathbf{79.2}$&$\mathbf{71.2}$&$\mathbf{61.1}$&$\mathbf{74.5}$&$\mathbf{69.7}$&$\mathbf{64.5}$&$\mathbf{71.8}$\\ \hline
			\end{tabular}
		}
		\vspace{0.1 cm}
		\caption{Comparisons with the state-of-the-art methods on multi-person pose tracking on ICCV'17 PoseTrack Challenge Dataset. \label{tab:test_tracking}}
		\vspace{-0.2 cm}
	\end{table}

	\begin{figure}
		\centering
		\includegraphics[width=0.45\textwidth]{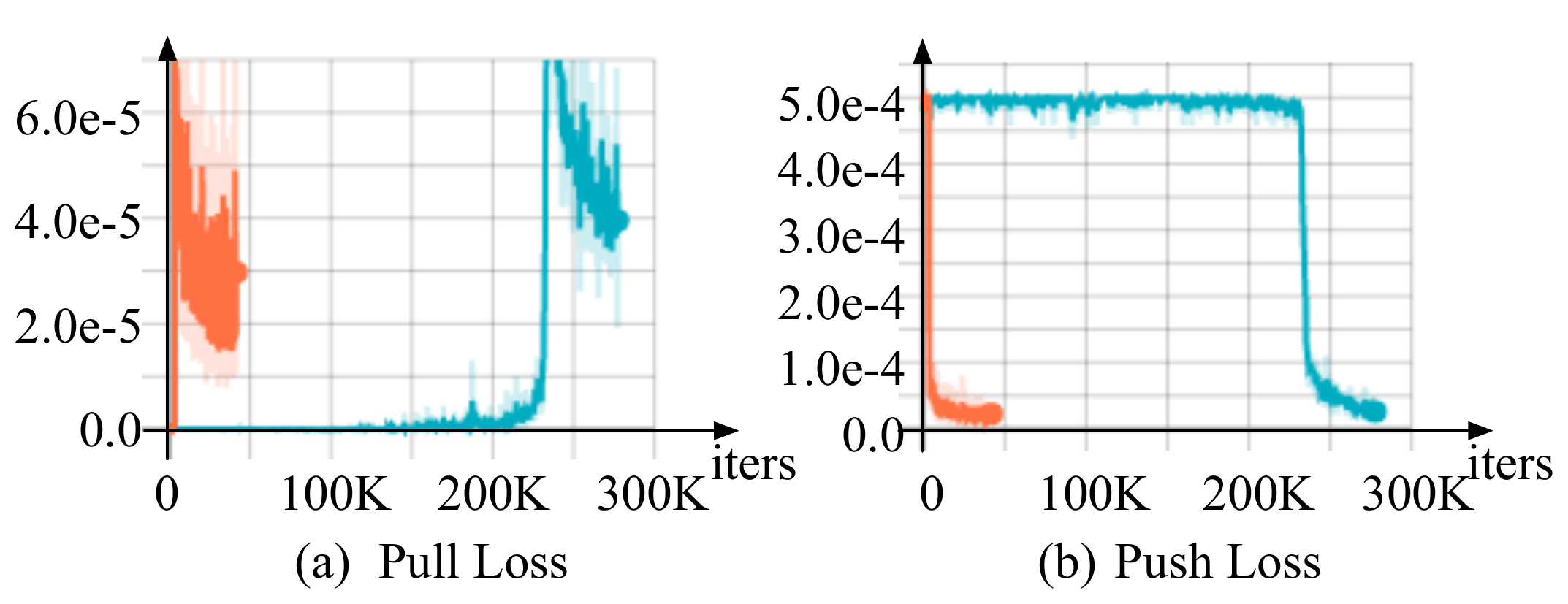}
		\caption{Learning curves of keypoint embedding (KE) with (orange) or without (cyan) auxiliary training.
		}
		\vspace{-0.4cm}
		\label{fig:learning_curve}
	\end{figure}

	\begin{figure}
		\centering
		\includegraphics[width=0.5\textwidth]{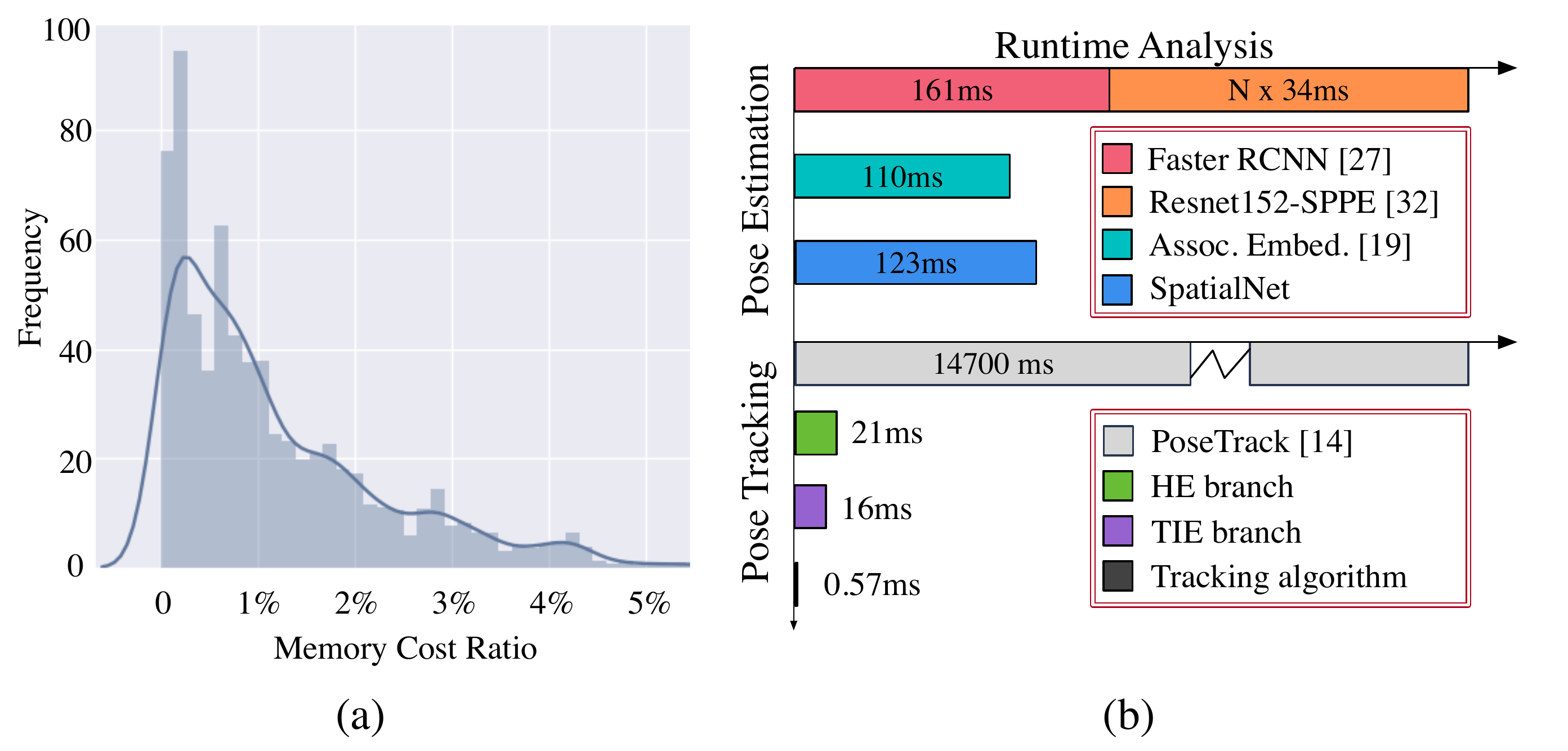}
		\caption{(a) Histogram of the memory cost ratio between PGG and GBMS~\cite{carreiraperpinan2008generalised} $\frac{\text{memory cost of PGG}}{\text{memory cost of GBMS}}$ on the PoseTrack val set. Using the instance-agnostic pose mask, PGG reduces the memory consumption to about $1\%$, \ie $100$ times more efficient. (b) Runtime analysis. CNN processing time is measured on one GTX-1060 GPU, while PoseTrack~\cite{Iqbal2016PoseTrack} and our tracking algorithm is tested on a single core of a 2.4GHz CPU. $N$ denotes the number of people in a frame, which is 5.97 on average for PoseTrack val set.}
		\vspace{-0.5cm}
		\label{fig:mem_time}
	\end{figure}
	\subsection{Ablation Study}
	
	We extensively evaluate the effect of each component in our framework. Table~\ref{tab:ICCV17} summarizes the single-frame pose estimation results, and Table~\ref{tab:ICCV17tracking} the pose tracking results. 
	
	For pose estimation we choose~\cite{newell2017associative} as our baseline, which proposes KE for spatial grouping. We also compare with one alternative embedding approach~\cite{liang2015proposal} for design justification. In \textbf{BBox}~\cite{liang2015proposal}, instance location information is encoded as the human bounding box (x, y, w, h) at each pixel. The predicted bounding boxes are then used to group keypoints into individuals. However, such representation is hard to learn due to large variations of its embedding space, resulting in worse pose estimation accuracy compared to \textbf{KE} and \textbf{SIE}. KE provides part-level appearance cues, while SIE encodes the human centroid constraints. When combined together, a large gain is obtained ($74.0$\% vs. $70.9$\%/$71.3$\%). As shown in Fig.~\ref{fig:learning_curve}, adding auxiliary tasks (\textbf{\emph{+aux}}) dramatically speeds up the training of KE, by enforcing geometric constraints on the embedding space. It also facilitates representation learning and marginally enhances pose estimation. As shown in Table~\ref{tab:ICCV17}, employing \textbf{PGG} significantly improves the pose estimation accuracy ($2.3\%$ for KE, $3.8\%$ for SIE, and $2.7\%$ for both combined). End-to-end model training and direct grouping supervision together account for the improvement. Additionally, using the instance-agnostic pose mask, the memory consumption is remarkably reduced to about $1\%$, as shown in Fig.~\ref{fig:mem_time}(a), demonstrating the efficiency of PGG. Combining both KE and SIE with PGG, further boosts the pose estimation performance to $77.0$\% mAP.
	
	For pose tracking, we first build a baseline tracker based on KE and/or SIE. It is assumed that KE and SIE change smoothly in consecutive frames, $\mathcal{K}(p^t_{j,k}) \approx \mathcal{K}(p^{t+1}_{j,k})$ and $\mathcal{S}(p^t_{j,k}) \approx \mathcal{S}(p^{t+1}_{j,k})$. Somewhat surprisingly, such a simple tracker already achieves competitive performance, thanks to the rich geometric information contained in KE and SIE. Employing TemporalNet for tracking significantly improves over the baseline tracker, because of the combination of the holistic appearance features of HE and temporal smoothness of TIE. Finally, incorporating spatial-temporal PGG to refine KE, SIE and TIE, further increase the tracking performance ($69.2$\% vs. $71.8$\% MOTA). We also compare with some widely used alternative tracking metrics, namely Object Keypoint Similarity (\textbf{OKS}), Intersection over Union (\textbf{IoU}) of persons and DeepMatching (\textbf{DM})~\cite{Revaud2015DeepMatching} for design justification. We find that TemporalNet significantly outperform other trackers with task-agnostic tracking metrics. OKS only uses keypoints for handling occlusion, while IOU and DM only consider human in handling fast motion. In comparison, we kill two birds with one stone.
	
	\textbf{MS-COCO Results.} Our SpatialNet substantially improves over our baseline~\cite{newell2017associative} on single frame pose estimation on the MS-COCO dataset. For fair comparisons, we use the same train/val split as~\cite{newell2017associative} for evaluation. Table~\ref{tab:coco_keypoint_results} reports both single-scale (\emph{sscale}) and multi-scale (\emph{mscale}) results. Four different scales $\{0.5, 1, 1.5, 2\}$ are used for multi-scale inference. Our \emph{sscale} SpatialNet already achieves competitive performance against \emph{mscale} baseline. By multi-scale inference, we further gain a significant improvement of 3\% AP. All reported results are obtained without model ensembling or pose refinement~\cite{cao2016realtime,newell2017associative}.
	
	\begin{table} \small
		\begin{center}
			\setlength{\tabcolsep}{3pt}
			\footnotesize
			\begin{tabular}{l|c|c|c|c|c|c|c|c} \hline
				& Head & Shou & Elb& Wri& Hip& Knee & Ankl & Total \\ \hline
				BBox~\cite{liang2015proposal} & $79.3$ & $75.6$ & $67.4$ & $60.2$ & $67.8$ & $61.6$ & $55.8$ & $67.7$ \\ \hline
				KE~\cite{newell2017associative} &$79.8$ & $77.7$ & $71.7$ & $63.4$ &  $71.4$ & $66.3$ &   $61.4$ &  $70.9$  \\ 
				SIE &$81.4$ & $78.8$ & $72.1$ & $64.2$ &  $72.2$ & $66.8$ &   $61.7$ &  $71.3$  \\ 
				KE+SIE &$82.2$ & $80.1$ & $74.7$ & $67.4$ &  $75.1$ & $69.4$ &   $64.6$ &  $74.0$  \\ 
				KE+SIE+\emph{aux} & $82.3$ & $80.3$ & $74.9$ & $67.8$ & $75.2$ & $70.1$ & $65.6$ & $74.3$ \\
				KE+PGG &$81.5$ & $80.0$ & $74.0$ & $65.8$ &  $73.4$ & $68.3$ &   $65.0$ &  $73.2$  \\ 
				SIE+PGG &$83.4$ & $80.6$ & $74.3$ & $67.4$ &  $76.0$ & $71.8$ &   $67.6$ &  $75.1$  \\ 
				Ours & $\mathbf{83.8}$ & $\mathbf{81.6}$ & $\mathbf{77.1}$ & $\mathbf{70.0}$ & $\mathbf{77.4}$ & $\mathbf{74.5}$ & $\mathbf{70.8}$ & $\mathbf{77.0}$ \\ \hline
			\end{tabular}
		\end{center}
		\caption{Ablation study on single-frame pose estimation (AP) on ICCV'17 PoseTrack validation set. \emph{aux} means auxiliary training with geometric ordinal prediction. Ours (KE+SIE+\emph{aux}+PGG) combines KE+SIE+\emph{aux} with PGG for accurate pose estimation.
			\label{tab:ICCV17}}
		\vspace{-0.3cm}
	\end{table}
	
	\begin{table} \small
		\begin{center}
			\setlength{\tabcolsep}{3pt}
			\footnotesize
			\scalebox{0.9}{
				\begin{tabular}{l|c|c|c|c|c|c|c|c} \hline
					& MOTA & MOTA & MOTA & MOTA & MOTA & MOTA & MOTA & MOTA \\
					& Head & Shou & Elb  & Wri  & Hip  & Knee & Ankl & Total\\ \hline
					OKS & $60.1$ & $60.4$ & $54.5$ & $47.1$ & $58.4$ & $57.0$ & $53.7$ & $56.2$  \\
					IOU & $62.5$ & $63.6$ & $54.3$ & $45.5$ & $59.3$ & $53.6$ & $48.6$ & $55.8$ \\
					DM~\cite{Revaud2015DeepMatching} & $62.9$ & $64.0$ & $54.6$ & $45.7$ & $59.6$ & $53.8$ & $48.7$ & $56.1$ \\ \hline
					KE & $72.9$ & $73.3$ & $64.6$ & $55.0$ & $68.7$ & $63.0$ & $58.5$ & $65.7$  \\
					KE+SIE & $75.4$ & $76.1$ & $67.0$ & $57.1$ & $70.9$ & $64.4$ & $59.4$ & $67.7$  \\ 
					HE & $76.0$ & $76.4$ & $67.7$ & $58.1$ & $71.7$ & $65.4$ & $60.5$ & $68.5$  \\
					TIE & $76.2$ & $76.7$ & $67.8$ & $58.4$ & $71.6$ & $65.3$ & $60.4$ & $68.6$  \\
					HE+TIE  & $76.9$ & $77.2$ & $68.4$ & $58.6$ & $72.4$ & $66.0$ & $61.2$ & $69.2$  \\
					Ours & $\mathbf{78.7}$ & $\mathbf{79.2}$ & $\mathbf{71.2}$ & $\mathbf{61.1}$ & $\mathbf{74.5}$ & $\mathbf{69.7}$ & $\mathbf{64.5}$ & $\mathbf{71.8}$\\ \hline
			\end{tabular}}
		\end{center}
		\vspace{-0.2cm}
		\caption{Ablation study on multi-person articulated tracking on ICCV'17 PoseTrack validation set. Ours (HE+TIE+PGG) combines HE+TIE with PGG grouping for robust tracking.
			\label{tab:ICCV17tracking}}
		\vspace{-0.2cm}
	\end{table}

	\begin{table}[!htb]\footnotesize
		\centering
		\scalebox{0.85}{
			\begin{tabular}{l|ccccc}
				&$AP$ & $AP^{.50}$ & $AP^{.75}$ & $AP^M$  & $AP^L$\\
				\hline
				Assoc. Embed.~\cite{newell2017associative} (\emph{sscale})& $0.592$ & $0.816$ & $0.646$ & $0.505$ & $0.725$  \\
				Assoc. Embed.~\cite{newell2017associative} (\emph{mscale}) & $0.654$ & $0.854$ & $0.714$ & $0.601$ & $0.735$  \\
				Ours  (\emph{sscale}) & $0.650$ & $0.865$ & $0.714$ & $0.570$ & $\mathbf{0.781}$  \\
				Ours  (\emph{mscale}) & $\mathbf{0.680}$ & $\mathbf{0.878}$ & $\mathbf{0.747}$ & $\mathbf{0.626}$ & $0.761$ \\ \hline
		\end{tabular}}
		\vspace{0.1cm}
		\caption{Multi-human pose estimation performance on the subset of MS-COCO dataset. \emph{mscale} means multi-scale testing.}
		\vspace{-0.5cm}
		\label{tab:coco_keypoint_results}
	\end{table}
	
	\subsection{Runtime Analysis}
	\label{sec:runtime}
	
	Fig.~\ref{fig:mem_time}(b) analyzes the runtime performance of pose estimation and tracking. For pose estimation, we compare with both top-down and bottom-up~\cite{newell2017associative} approaches. The top-down pose estimator uses Faster RCNN~\cite{renNIPS15fasterrcnn} and a ResNet-152~\cite{he2016deep} based single person pose estimator (SPPE)~\cite{xiao2018simple}. Since it estimates pose for each person independently, the runtime grows proportionally to the number of people.
	
	Compared with~\cite{newell2017associative}, our SpatialNet significantly improves the pose estimation accuracy with the increase of limited computational complexity. For pose tracking, we compare with the graph-cut based tracker (PoseTrack~\cite{Iqbal2016PoseTrack}) and show the efficiency of TemporalNet.
	
	\section{Conclusion}
	\label{sec:conclusion}
	We have presented a unified pose estimation and tracking framework, which is composed of SpatialNet and TemporalNet: SpatialNet tackles body part detection and part-level spatial grouping, while TemporalNet accomplishes the temporal grouping of human instances. We propose to extend KE and SIE in still images to HE appearance features and TIE temporally consistent geometric features in videos for robust online tracking. An effective and efficient Pose-Guided Grouping module is proposed to gain the benefits of full end-to-end learning of pose estimation and tracking.
	
	{\small
		\bibliographystyle{ieee}
		\bibliography{egbib}
	}
	
\end{document}